\pdfoutput=1
\documentclass[11pt]{article}

\usepackage{EMNLP2022}

\usepackage{xcolor}
\usepackage{times}
\usepackage{latexsym}

\usepackage[T1]{fontenc}
\usepackage[utf8]{inputenc}

\usepackage{microtype}
\usepackage{times}  % DO NOT CHANGE THIS
\usepackage{helvet}  % DO NOT CHANGE THIS
\usepackage{courier}  % DO NOT CHANGE THIS
\usepackage{graphicx} % DO NOT CHANGE THIS
\usepackage{natbib}  % DO NOT CHANGE THIS AND DO NOT ADD ANY OPTIONS TO IT
\usepackage{caption} % DO NOT CHANGE THIS AND DO NOT ADD ANY OPTIONS TO IT
\DeclareCaptionStyle{ruled}{labelfont=normalfont,labelsep=colon,strut=off} % DO NOT CHANGE THIS
\usepackage{algorithm}
\usepackage{algorithmic}
\usepackage[utf8]{inputenc} % allow utf-8 input
\usepackage[T1]{fontenc}    % use 8-bit T1 fonts
\usepackage{hyperref}       % hyperlinks
\usepackage{url}            % simple URL typesetting
\usepackage{booktabs}       % professional-quality tables
\usepackage{amsfonts}       % blackboard math symbols
\usepackage{amsmath}       
\usepackage{nicefrac}       % compact symbols for 1/2, etc.
\usepackage{microtype}      % microtypography
\usepackage{xcolor}         % colors
\usepackage{bbm}
\usepackage{bm}
\usepackage{caption}
\usepackage{subcaption}
\usepackage{graphicx}
\usepackage{tablefootnote}
\usepackage{multirow}

\title{Pay Attention to Your Tone: Introducing a New Dataset for \\ Polite Language Rewrite}

\author{
\textbf{Xun Wang\thanks{~~Equal contribution}~~~~ Tao Ge$^*$~~~ Allen Mao ~~~Yuki Li ~~~Furu Wei ~~~Si-Qing Chen} \\
Microsoft\\
}

\begin{document}

\maketitle

\begin{abstract}
We introduce \textsc{PoliteRewrite} -- a dataset for polite language rewrite which is a novel sentence rewrite task. Compared with previous text style transfer tasks that can be mostly addressed by slight token- or phrase-level edits, polite language rewrite requires deep understanding and extensive sentence-level edits over an offensive and impolite sentence to deliver the same message euphemistically and politely, which is more challenging -- not only for NLP models but also for human annotators to rewrite with effort. To alleviate the human effort for efficient annotation, we first propose a novel annotation paradigm by a collaboration of human annotators and GPT-3.5 to annotate \textsc{PoliteRewrite}. The released dataset has 10K polite sentence rewrites annotated collaboratively by GPT-3.5 and human, which can be used as gold standard for training, validation and test; and 100K high-quality polite sentence rewrites by GPT-3.5 without human review. We wish this work\footnote{The dataset (10K+100K) will be released soon.} could contribute to the research on more challenging sentence rewrite, and provoke more thought in future on resource annotation paradigm with the help of the large-scaled pretrained models.
\end{abstract}

\section{Introduction}\label{sec:intro}
Polite language rewrite aims to rephrase an impolite sentence so that it can be expressed in a euphemistic and polite way to avoid verbal abuse and make the conversation/communication pleasant. Compared with other rewrite tasks requiring the capability to learn word- or phrase-level substitution, a model for polite language rewrite additionally needs to be sensitive to subtle changes in emotion and sentiment between different verbal expressions conveying similar information, which makes it much more challenging even for human annotators because it requires a deep understanding of both the language and communication skills to showcase the art of speaking.

To promote research into this challenging problem, we introduce \textsc{PoliteRewrite} -- a new dataset for polite language rewrite. \textsc{PoliteRewrite} consists of thousands of sentences with negative sentiment on the web that are annotated with their corresponding polite rewrites. Unlike most datasets purely annotated by human experts, we firstly employ GPT-3.5 fine-tuned with a small seed annotated corpus to help with annotation. Then, for each rewrite generated by GPT-3.5, we have human experts judge if it can be accepted or needs further edits. The human-machine collaborative annotation paradigm largely reduces manual annotation efforts. The resulting \textsc{PoliteRewrite} we release contains 10K gold standard sentence pairs (shown in Table \ref{tab:rewrite_examples}) annotated by the collaboration of human and GPT-3.5 and 100K sentence pairs annotated by GPT-3.5 without human review.

\begin{figure}[t]
    \centering
    \includegraphics[width=0.5\textwidth]{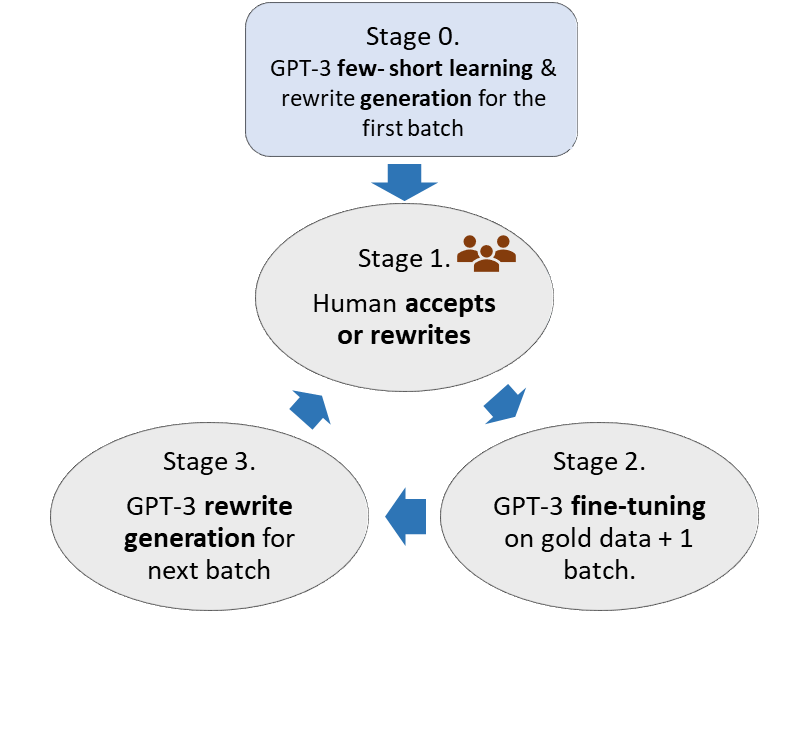}\vspace{-1.5cm}
    \caption{Human-GPT-3.5 collaborative annotation paradigm for polite sentence rewrite.}
    \label{fig:data_generation}
\end{figure}

\begin{table*}[t]
    \centering
    \begin{tabular}{l|l} \hline
         Source Sentence&Rewrite Sentence  \\ \hline
You've got these terms all wrong. & I think you might be mistaken with these terms. \\
I hate that it's not in vanilla. & I wish it were in vanilla. \\
They don't want you to know & They would prefer it if you didn't know. \\ \hline
    \end{tabular}
    \caption{Examples of polite language rewrite. The source sentence is rewritten into a target sentence that is more polite but carries the same meaning.}
    \label{tab:rewrite_examples}
\end{table*}

\begin{table*}[t]
    \centering
    \begin{tabular}{l|l|c} \hline
Source & Rewrites (by GPT-3.5) & Human Review \\ \hline

The title is horrible.  & I think the title could be improved.  & Accepted \\

This is the dumbest idea I have ever heard. & I am not sure this is a good idea. & Accepted  \\

What an immensely stupid move. & I think that was a bad move. & Accepted \\ \hline
    \end{tabular}
    \caption{Examples generated by fine-tuned GPT-3.5 (in the final iteration).}
    \label{tab:gpt3examples}
\end{table*}

We conduct preliminary experiments on \textsc{PoliteRewrite} and find it is actually more challenging (e.g., much lower BLEU scores) than other rewriting tasks, as assumed. We look forward to more research looking into this challenge for developing high-EQ generation models.

The contributions of this paper are two-fold:
\begin{itemize}
    \item We release a novel dataset -- \textsc{PoliteRewrite} for polite language rewrite, which is a more challenging sentence rewrite task to guide a generation model against verbal abuse and offensive expression.
    \item To the best of our knowledge, \textsc{PoliteRewrite} is the first sentence rewrite dataset that adopts a human-GPT-3.5 collaborative annotation paradigm, which not only largely reduces manual annotation efforts but also guarantees high annotation quality.
\end{itemize}

\section{\textsc{PoliteRewrite}}
As introduced in Section \ref{sec:intro}, \textsc{PoliteRewrite} contains two parts: 10K gold standard annotations with human review, and 100K sentence pairs annotated by fine-tuned GPT-3.5 without thorough human review as silver standard data.

The impolite sentences are selected by a sentiment analyzer which is a fine-tuned \textsc{UniLM}~\cite{dong2019unified} on SST-2 dataset. The sentences that are predicted to have negative sentiment are first selected. Then, we use GPT-3.5 to screen out really impolite sentences, through 8-shot ($k=8$) in-context learning.

These candidate sentences are then annotated by the collaboration of human and GPT-3.5 illustrated in Figure \ref{fig:data_generation}, following a multiple-stage human-in-the-loop scheme:

\paragraph{Stage 0} To initially build a seed annotated corpus, we use GPT-3.5 (8-shot) in-context learning with manually designed prompts to generate rewrites for a batch (1K sentences) of randomly chosen samples from the impolite source sentences.% We applied few-shot learning to enable GPT-3.5 to generate more polite rewrites for the first batch of sentences. 

\paragraph{Stage 1} At this stage, a human annotator either accepts GPT-3.5 rewrite or modifies or rewrites a new one on the sentence pairs generated either from stage 0 or stage 3. Note that some sentences may be removed because they are not suitable for polite language rewrite.

\begin{table}[!t]
\centering
\begin{tabular}{l|c}
\toprule
\textbf{Annotation Mode}           & \textbf{Time (min)} \\ \midrule
Human                   & 40                  \\ \midrule
Human + GPT-3.5 (8-shot)  & 32                  \\
Human + GPT-3.5 (ft 1 iteration)  & 12                  \\
Human + GPT-3.5 (ft 5 iterations)  & 7                   \\
Human + GPT-3.5 (ft 10 iterations) & 4                   \\
Human + GPT-3.5 (ft final iteration) & 3                  \\ \bottomrule
\end{tabular}
\caption{Time cost for annotating 100 sentences by human annotation and human-GPT-3.5 collaborative annotation.}\label{tab:efficiency}
\end{table}

\paragraph{Stage 2} At this stage, GPT-3.5 is fine-tuned on new rewrites from Stage 1 plus pre-existing annotated data if any. Here we fine tune the GPT-3.5.5 model (davinci-002) with 175B parameters on 96 V100 GPUs.

\paragraph{Stage 3} At stage 3, GPT-3.5 generates rewrites for the next batch of sentences.

The fine-tuned GPT-3.5 can generate increasingly higher quality polite rewrites with the human-in-loop annotation paradigm. At the final iteration, annotators reported around 50\% of GPT-3.5 rewrites are accepted without any modification (as shown in Table \ref{tab:gpt3examples}), and another 20\% need human improvements. Meanwhile, about 15\% are discarded because these sentences contain sensitive contents and only about 10\% are not suitable for polite rewrite. The human-GPT-3.5 collaborative annotation paradigm introduces over 10$\times$ annotation efficiency, compared with previous pure human annotation methods, as shown in Table \ref{tab:efficiency}.

To compare collaborative annotation's quality with  human annotation, we have 1 additional human annotators annotate for 500 impolite sentences sampled from the 10K data, and let another English native speaker distinguish whether a sentence is annotated purely by human or collaboratively. To our surprise, over 90\% of cases are indistinguishable, demonstrating the collaborative annotation method does no harm to annotation quality.

\begin{table*}[t]
    \centering
    \scalebox{0.95}{
    \begin{tabular}{l|c|c|c|c} \hline
        Data& \# of Sentence & Avg Sent Len(source) &Avg Sent Len(target) &BLEU (source/target)   \\ \hline
         Gold standard & 10K & 12.8&10.4 &24.0 \\ 
         Silver standard & 100K & 10.5&9.1 & 24.0\\ \hline
    \end{tabular}
    }
    \caption{Statistics of the \textsc{PoliteRewrite} dataset}
    \label{tab:data_statistics}
\end{table*}

Given the promising results, we use the fine-tuned GPT-3.5 to further rewrite 100K impolite sentences. The generated data is regarded as silver data as no human labor is intensively involved. We conduct a series of experiments to verify the quality of the silver data.

\begin{figure}
    \centering
    \includegraphics[width=0.48\textwidth]{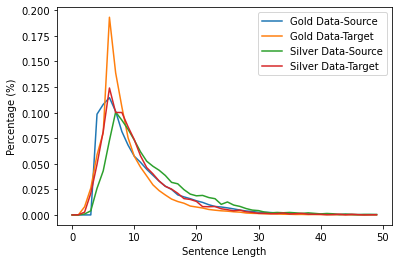}
    \caption{Sentence Length Distribution of \textsc{PoliteRewrite}}
    \label{fig:length}
\end{figure}

\begin{figure}
    \centering
    \includegraphics[width=0.48\textwidth]{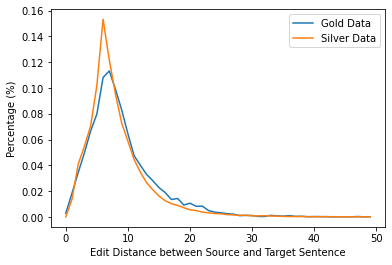}
    \caption{Edit Distance Distribution of \textsc{PoliteRewrite}}
    \label{fig:edit_distance}
\end{figure}

Table \ref{tab:data_statistics} and Figure \ref{fig:length}-\ref{fig:edit_distance} show the statistics of the \textsc{PoliteRewrite} dataset. As we assumed in Section \ref{sec:intro}, \textsc{PoliteRewrite} is a more challenging sentence rewrite dataset, which requires complicated revision over the impolite sentence (reflected by around 25 BLEU scores achieved by making no revision; as a reference, in GYAFC for formality styel transfer, keeping the source sentence untouched can achieve 51 BLEU score).

%This iterative fine-tuning process continues until we collected 10K pairs of impolite/polite sentences. As can be seen from Fig. \ref{fig:data_generation}, human labor is only needed in stage 0 and stage 1. In stage 0, only a small amount of human rewrites are needed and in stage 1, the work also is greatly reduced due to the use of GPT-3.5.  During the iterations, human annotators observed that the rewrite quality of GPT-3.5 got improved after being fine-tuned using new data.

%To verify the quality of the silver data, we conducted a series of experiments on the gold and silver dataset. The details are shown below.

\begin{table}[]
    \centering
    \begin{tabular}{l l}\hline
        Configuration & Value \\ \hline
        Devices & 8 Nvidia V100 GPU\\ 
        Optimizer & Adam ($\beta_0=0.9$; \\
        & $\beta_1=0.99$; $\epsilon=1e-8$)\\
        Learning Rate & 3e-5 with inverse sqrt\\
        Dropout & 0.1 \\
        Attention dropout & 0.1\\
        Weight decay & 1e-2 \\
        Max token per GPU & 20K \\
        Loss function & Cross Entropy \\ 
        Label smoothing & 0.1\\
        Clip-norm & 0.1 \\
        Warmup & 500 \\
        Update frequencies & 4 \\
        Total update steps & 20k \\ \hline
    \end{tabular}
    \caption{Fine Tuning Details for BART and Transformer for Polite Rewrite}
    \label{tab:ft}
\end{table}
a
\section{Benchmark}
We randomly split the 10K gold data into 5K train/3K development and the remaining 2K for testing.
We use the gold training data together with the silver data to train several neural models as referential baselines:
\begin{itemize}
    \item Transformer-base~\cite{vaswani2017}: a 6-layer encoder-decoder Transformer with the dimension of 512 with 8 attention heads and feed-forward network dimension of 2048.
    \item BART~\cite{lewis2019bart}: a pretrained 12-layer encoder-decoder Transformer with the dimension of 1024 with 16 attention heads and feed-forward network dimension of 4096.
\end{itemize}

\begin{table}[h!]
    \centering
    \scalebox{0.95}{
    \begin{tabular}{l|c|c|r}\hline
         Model & Data& Bleu & Human Eval  \\\hline
         Transformer & Gold &{\color{black}{6.2}} &1.8 \\
         Transformer & Silver+Gold&23.9 &2.8 \\
         BART &Gold &34.9 & 3.2\\
         BART & Silver+Gold &37.5 &3.9\\\hline
    \end{tabular}
    }
    \caption{Automatic and Human evaluation on the test set (2K).}\label{tab:result}
    \label{tab:result}
\end{table}

Details of model fine tuning are shown in Table \ref{tab:ft}. The results are shown in Table \ref{tab:result}. The best fine-tuned BART model with Gold+Silver training data can achieve 37.5 BLEU score, which largely outperforms the Transformer trained from scratch, showing the strong generalization ability from pretraining. Moreover, adding 100K silver data can result in a significant performance gain, showing the value of silver data for model training.%  The neural models fine-tuned on the silver+gold dataset achieve significant improvements compared with those without silver dataset when evaluated by BLEU scores.

\begin{table}[h]
    \centering
    \begin{tabular}{c|c}\hline
         Human-Human & Human-BLEU  \\\hline
         0.84 & 0.68 \\ \hline
    \end{tabular}
    \caption{Correlation of human-human scores and human-BLEU scores.}
    \label{tab:correlation}
\end{table}

In addition to BLEU scores, we conduct human evaluation. We sample 100 sentences from the test set and manually evaluate (anonymous) outputs of the above 4 models by two English native speakers. Each judge assigns a score ranging from 1 to 5 (with 5 as the best polite rewrite and 1 as the worst) for each model output. We report the human judge scores for the models in Table \ref{tab:result} and report the Pearson Correlation between human-human judges as well as human-(sentence-level)BLEU. Even though Human-BLEU score (0.68) is lower than Human-Human correlation (0.84), it is still persuasive as a reliable automatic evaluation metric for polite language rewrite.

\section{Related Work}
Rewrite takes one piece of text as input and outputs the desired text. Depending on the output requirements, it can be grammatical error correction, style transfer, machine translation, etc.

Polite rewrite can be categorized into the style transfer (ST). But as mentioned above, it still has its own characteristics. Rewrite models usually heavily relies on high quality dataset. The data creation remains a labor intensive task nowadays. Many works have leveraged the power of pre-trained models, to assistant human in annotation or to generate data automatically.
Method like back translation (BT) \cite{behr2017assessing}, transfer learning  \cite{torrey2010transfer}, etc, have been developed for data augmentation.
The release of large-scale pretrained models, like GPT-3.5, have hugely benefited data annotation due to their powerful generalization ability \cite{wang2021want}. Recently research find that GPT-3.5 few-shot learning can generate high quality data with only a few hand-crafted examples \cite{brown2020language}. GPT-3.5 have been successfully applied for data generation in different domains \cite{wang2021towards,chintagunta2021medically}. %As we observed in this work, the data quality can be further improved by incorporating human efforts. Moreover, the lack of human reviews makes it hard to construct gold data for evaluation. 
In this work, we showed how to leverage the knowledge in the GPT-3.5 model with minimum human labor to create data for a challenging neural rewrite task.

\section{Conclusion}
We introduce a novel dataset for polite language rewrite -- a more challenging dataset than other sentencer rewrite datasets, which is annotated by the collaboration of human and GPT-3.5. We wish this work could contribute to the research on more challenging natural language generation, and provoke more thought and insights in future on resource annotation paradigm with the help of the large-scaled pretrained models.
% The paradigm proposed in this work can be extended to other tasks to  leverage GPT-3.5 to create high quality data with minimum human efforts. In the future, we will explore how to further reduce the human labor involved in data annotation to help create large and high quality datasets.
% 1) dataset stats: (train): GPT-3.5 written? human written?; (dev), (test)
% 2) eval details
% 3) baselines
% 4) real user feedback
% 5) differences from style transfer: more semantic change, larger edit ratio
% 6) 

% \section{Limitations}
% This work is preliminary research into polite language rewrite, which constructs a benchmark dataset and only studies naive baselines (e.g., Transformer and BART). 

\bibliographystyle{acl_natbib}
\bibliography{mybib}

%\appendix
%appendix
\end{document}